%% file: paper.tex
\documentclass[review]{elsarticle}

\usepackage{lineno,hyperref}

\usepackage{graphicx}
\usepackage{subfigure}
\usepackage{bm}
\usepackage[cmex10]{amsmath}
\usepackage{algorithmic}
\usepackage[ruled]{algorithm2e}
\usepackage{booktabs}
\usepackage{multirow}
\usepackage{color}
\usepackage{makecell}
\usepackage{amssymb}
\usepackage{subfiles}
\usepackage{floatrow}
\modulolinenumbers[5]

\journal{Pattern Recognition}

\newcommand{\FIG}[1]{Figure \ref{#1}}
\newcommand{\TABLE}[1]{Table \ref{#1}}

\newcommand{\MODEL}{CSCNet}

\begin{document}

\begin{frontmatter}

\title{CSCNet: Contextual Semantic Consistency Network for Trajectory Prediction in Crowded Spaces}

\author{Beihao Xia}
\ead{xbh_hust@hust.edu.cn}
\author{Conghao Wong\fnmark[1]}
\ead{conghao_wong@icloud.com}
\author{Qinmu Peng}
\ead{pengqinmu@hust.edu.cn}
\author{Wei Yuan}
\ead{yuanwei@mail.hust.edu.cn}
\author{Xinge You\cormark[*]}

\cortext[*]{Corresponding author: youxg@hust.edu.cn}
\address{Huazhong University of Science and Technology, Luoyu Road 1037, Wuhan, China}
\fntext[1]{Equal contribution.}



\begin{abstract}
Trajectory prediction aims to predict the movement trend of the agents like pedestrians, bikers, vehicles.
It is helpful to analyze and understand human activities in crowded spaces and widely applied in many areas such as surveillance video analysis and autonomous driving systems.
Thanks to the success of deep learning, trajectory prediction has made significant progress.
The current methods are dedicated to studying the agents' future trajectories under the social interaction and the sceneries' physical constraints.
Moreover, how to deal with these factors still catches researchers' attention.
However, they ignore the \textbf{Semantic Shift Phenomenon} when modeling these interactions in various prediction sceneries.
There exist several kinds of semantic deviations inner or between social and physical interactions, which we call the ``\textbf{Gap}''.
In this paper, we propose a \textbf{C}ontextual \textbf{S}emantic \textbf{C}onsistency \textbf{Net}work (\textbf{CSCNet}) to predict agents' future activities with powerful and efficient context constraints.
We utilize a well-designed context-aware transfer to obtain the intermediate representations from the scene images and trajectories.
Then we eliminate the differences between social and physical interactions by aligning activity semantics and scene semantics to cross the Gap.
Experiments demonstrate that CSCNet performs better than most of the current methods quantitatively and qualitatively.
\end{abstract}

\begin{keyword}
\texttt{Trajectory Prediction \sep The Context-aware Transfer \sep The Conditional Context Loss }
\end{keyword}

\end{frontmatter}


\subfile{./content/Introduction.tex}

\subfile{./content/RelatedWork.tex}

\subfile{./content/Method.tex}

\subfile{./content/Experiments.tex}

\subfile{./content/Discussion.tex}

\subfile{./content/Conclusion.tex}

\newpage

\bibliography{myref}



\end{document}

%% file: content/Introduction.tex
\section{Introduction}

\subfile{fig_task.tex}

It is meaningful and challenging to model and understand agents' activities in a variety of applications, including self-driving \cite{desire,rhinehart2018r2p2,rhinehart2019precog}, robotic navigation \cite{unfreezing}, tracking \cite{learningSocialEtiquette}, and so on.
Trajectory prediction would like to predict the agents' future trajectories considering the agent-agent influences \cite{socialForce,socialLSTM} and the sceneries' physical constraints \cite{ssLSTM,sophie,bigat}, as shown in \FIG{fig_task}.
Furthermore, it can be regarded as an essential component to recognize and analyze agents' behaviors \cite{deo2018would,youWillNeverWalkAlone}.
A large number of \cite{pei2019human,barata2021sparse} researchers are committed to this task and make their contributions.
However, trajectory prediction remains a considerable challenge due to agents' uncertain future expectations, variable physical scenarios, and complex social behaviors.

In crowded spaces, agents' short-term behaviors or activities may be easily influenced by frequent socially interactive behaviors and the surroundings.
Previous methods like \cite{rossi2021human,multiAgentTensorFusion} have classified these two factors that may affect agents' future activities into \textbf{Social Interaction} i.e., agent-agent interaction, and \textbf{Physical Interaction} i.e., agent-scene interaction.
Moreover, a lot of networks or modules \cite{huang2021lstm,zamboni2022108252} are employed to obtain the corresponding activity semantics and scene semantics, respectively.
For instance, SoPhie \cite{sophie}, and S-BiGAT \cite{bigat} use Convolutional Neural Networks (CNNs) to extract visual features of scene images (scene semantics) on one side, and design another attention-based social module to obtain the activity semantics on the other side.
Outstanding performance can be achieved with the joint efforts of these two different interactive cues.

However, they ignore the \textbf{Semantic Shift Phenomenon} when modeling these interactions in various prediction sceneries.
There exist several kinds of semantic deviations inner or between social and physical interactions, which we call the ``\textbf{Gap}''.
For example, the park's lawn allows for walking and resting, but few pedestrians attempt to do so on the lawn beside sidewalks.
This is a gap between the theoretical meaning and the actual behavior, which also exists in social and physical interactions.
Social interaction and physical interaction reflect the same thing from two different perspectives.
The depiction of two interactions are expected to work together to model agents' interactive behaviors.
However, most previous methods employ different forms of features to describe these interactive behaviors without any normalization or regulation operations.
Besides, few of them further align them semantically, leading to another kind of ``Gap'', which affects the performance of prediction models.
This makes it difficult for the two semantics to work as they should.

\subfile{./fig_introduction.tex}

As shown in \FIG{fig_introduction}, the mentioned ``Gap'' exists in the following aspects, which lead to the semantic shift phenomenon when modeling these interactive cues.
\begin{itemize}
    \item Physical Gap.
    Prediction scenarios change flexibly and dynamically.
    Factors like the density of traffic, the appearance and distribution of buildings vary widely.
    Meanwhile, even the same scene objects may play different roles in different prediction scenarios.
    These differences will make it difficult to model the physical interaction.
    \item Social Gap.
    On one side, different kinds of agents (e.g., adults, kids, bicyclists, vehicles.) have their own activity styles, leading to differences in interaction preferences.
    On the other side, the same kind of agents may also exhibit different interactive behaviors when facing different scenarios.
    For instance, students' behaviors and interaction patterns in school may differ from the corresponding patterns they perform in the park.
    Additionally, the scale differences and coordinate transformation relationships of trajectory data in different datasets also pose challenges for descriptions and analyses of social behavior across datasets.
    \item Semantic Gap.
    There are differences in the expression of scene constraints (primarily obtained in the form of images) and social constraints (primarily obtained in the form of coordinates).
    Data from different modalities will make analysis more difficult.
\end{itemize}

In this work, we aim to investigate the differences in interactive behaviors and semantic differences.
Besides, we try to eliminate the differences through a specific transfer network to achieve effective contextual modeling for the future activities of agents.
Therefore, we propose a contextual semantic consistency network (CSCNet) to cross these gaps.
First, we use a context-aware transfer sub-network to obtain the intermediate representations describing social and physical interactions.
Second, we transfer these data with distribution shifts and different manifestations to a common feature space to eliminate semantic differences.
Moreover, the multiple modules are trained end-to-end using a multi-target loss function that penalizes average error, intermediate representation accuracy, and prediction consistency with the generated prior.
After that, we will predict agents' future trajectories under the corresponding priors to make predictions conforming to social rules and physical constraints.

The main contributions of this work are listed as follows:
\begin{itemize}
    \item We design a context-aware transfer to break the ``Gap'' within social and physical descriptions by aligning activity and scene semantics.
    \item We use a novel conditional context loss to train the whole network end-to-end to make predictions in line with social and physical rules.
    \item Combining the context-aware transfer and conditional context loss, CSCNet outperforms the existing models on ETH-UCY and SDD Datasets.
\end{itemize}

The rest of the paper is organized as follows:
We give a brief overview of related work in Section 2.
In Section 3, we will describe our model in detail.
We show the experimental analysis of CSCNet in Section 4.
We also discuss the limitations of CSCNet in Section 5.

%% file: content/fig_task.tex
\begin{figure*}[t]
\centering
\includegraphics[scale=0.2]{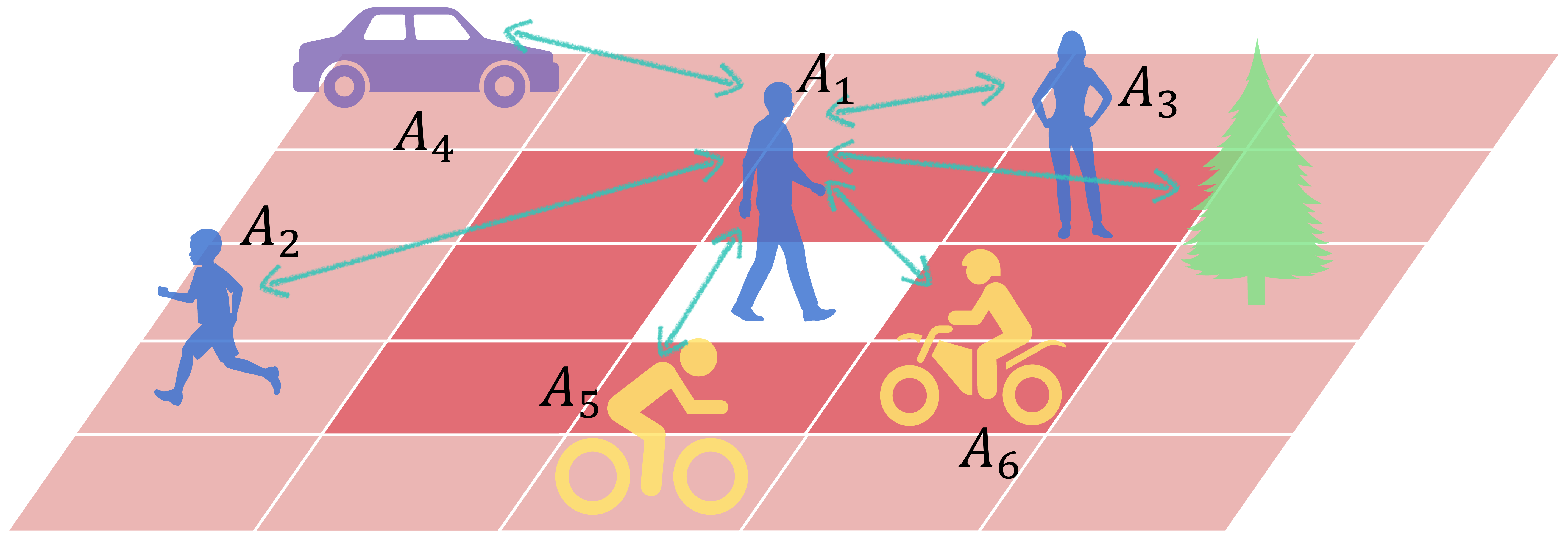}
\caption{
When predicting the future trajectory of the male $A_1$, we should take the interactive context into consideration.
Different kinds of agents, containing the kid $A_2$, the female $A_3$, the vehicle $A_4$, the bicycler $A_5$, and the motorcyclist $A_6$, would affect his movement and future decisions.
Scene interactive objects, like the trees, might also bring physical constraints.
}
\label{fig_task}
\end{figure*}

%% file: content/fig_introduction.tex
\begin{figure*}[t]
\centering
\includegraphics[width=1\columnwidth]{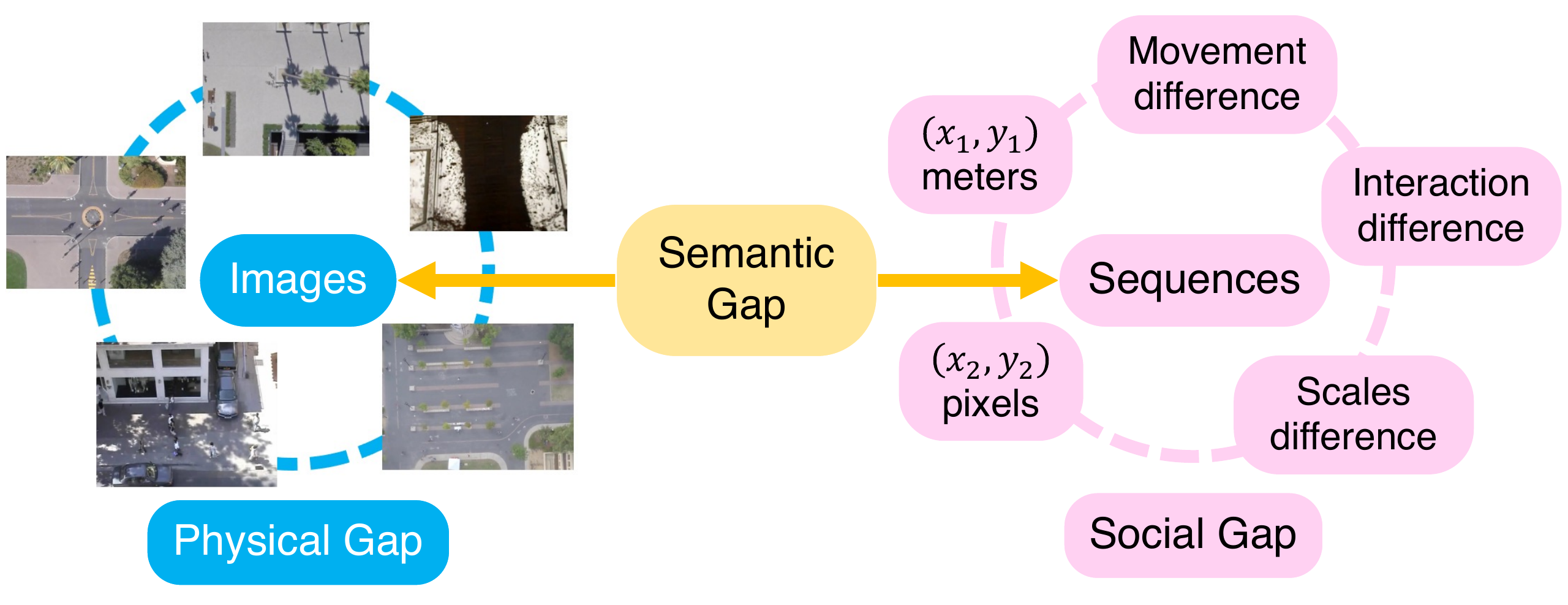}
\caption{
We focus on crossing the ``Gap'', i.e, addressing the semantic shift phenomenon in modeling interactions. 
It contains Physical Gap, Social Gap, and Semantic Gap.
}
\label{fig_introduction}
\end{figure*}

%% file: content/RelatedWork.tex
\section{Related Work}
Trajectory prediction has made significant progress thanks to many researchers' continuous research in the community.
They continuously study four main aspects: the modeling of social interaction, the description of environmental physical constraints, the joint representation of possible cues, and multi-modal predictions.
In this work, we focus on the first three: learning the intermediate representations for the physical cues and social cues with the same structured representations.
Then we will introduce the relevant works in the following.

\subsection{Social Interaction in Trajectory Prediction}
Traditionally, some classical works \cite{socialForce,activityForecasting,unfreezing} describe social interaction with hand-crafted features, like speed, acceleration, and other social force items \cite{youWillNeverWalkAlone}.
Despite the improvement, these methods model interactions by manually adjusting parameters rather than inferring them in a data-driven fashion.
A large number of data-driven methods have been proposed due to the rapid development of deep learning.
As the pioneering work, S-LSTM \cite{socialLSTM} introduces a ``Social Pooling'' module to adaptively learn typical interactions.
This pooling layer allows spatially proximal LSTMs to share information with each other, thus achieving better performance.
However, S-LSTM does not differentiate the effect of neighboring pedestrians.
Hence, Xu et al. \cite{cidnn} use the spatial affinity to measure the degrees of influences more precisely.
Similarly, Vemula et al. \cite{socialAttention} adopt a soft attention model over all humans in the crowd to avoid the local neighborhood assumption.

Additionally, more new modules are designed to deal with social interactions, i.e., the max-pooling module \cite{socialGAN}, the sort-pooling module \cite{sophie}.
Another line of work uses Graph Neural Networks (GNNs) to build excellent relational representations.
A series of GNN-based models have demonstrated their superiority in modeling social interaction, e.g. a spatio-temporal graph network in \cite{socialAttention}, a spatio-temporal graph attention network in \cite{stgat}, a heterogeneous graph network in \cite{trafficPredict}, a spatio-temporal graph convolutional neural network in \cite{stgcnn}.
Most of these deep-learning-based methods ignore the possible differences in interactive preferences in different prediction scenarios.
In contrast, we propose a method to describe the interactive behaviors in a more generic energy fashion.

\subsection{Environmental Physical Constraints}
Environmental physical constraints can restrict agents' activity areas and thus play almost the same role as social interaction.
A large body of work has investigated how agents' plannings can be affected by the environment to improve prediction performance via scene images.

Robicquet et al. \cite{learningSocialEtiquette} annotate scenes with labels like road, roundabout, sidewalk, grass, building, and bike rack to make sure which areas are available for agents' activities.
Although this helps somewhat, these labels are static and coarse for agents to make future decisions.
Regardless of the types of labels, agents only care about if the areas can be active.
Some recent works \cite{ssLSTM,sophie,bigat} have adopted a different mindset.
Sadeghian et al. in \cite{sophie} first use a Convolutional Neural Network (CNN) to extract the visual features from the images to get scene semantics.
Despite the automatic label generation, the above methods have the same limitations as manual labels \cite{learningSocialEtiquette}.
However, different agents can exhibit various behaviors in similar scenarios, not to mention in flexible and dynamic ones.
These methods rarely consider the mapping relationship between scenes and agents' activities.
Besides, agents' interactive behaviors can be affected by scene context dynamically.

Another type of method \cite{peekingIntoTheFuture,Liang_2020_CVPR} takes advantage of semantic segmentation to label the scene objects instead of manual annotation.
Liang et al. \cite{peekingIntoTheFuture} first use a pre-trained scene segmentation model to extract pixel-level scene semantic classes, which contain totally 10 common scene classes, like roads, sidewalks, etc.
Agents always synthesize and analyze scene information within a certain range of areas before making decisions.
The use of segmentation maps is not sufficient to fully reflect their future choices.
In this work, we aim to exclude differences when modeling physical interactions in multiple complex scenes.

\subsection{Social and Physical Interactions Fusion}
There exists a gap within social and physical descriptions since they describe factors that affect agents' behaviors from two entirely different aspects.
For instance, the methods \cite{sophie,multiAgentTensorFusion} extract the visual features from the images to get scene semantics.
Meanwhile, they extract the behavioral features from the historical trajectories to obtain activity semantics.
Two semantics are naturally fused with a simple concatenation without any other consideration.
Most current methods model the two cues together through an attention mechanism \cite{sophie}, which makes it difficult to fully consider the joint effects due to the absence of any regulation, normalization, and further semantic level of alignment.
Additionally, more detailed extra information is available in recent works like the bounding boxes of the objects and activity labels in \cite{peekingIntoTheFuture}, the group annotations by experts in \cite{Sun_2020_CVPR}, to improve the performance.
Although these methods attempt to jointly characterize agents' activities through inputs from different modalities, most of them do not align these features on semantics.
That is, there is a semantic shift phenomenon when modeling potential influencing factors in the above methods.
In this paper, we would eliminate the semantic differences by modeling the social and physical cues into the common feature space and then aligning them at the semantic level.

%% file: content/Method.tex
\section{Model}

\subfile{./fig_model.tex}

\subsection{Problem Formulation}
Given a set of $N$ agents' historical trajectories $\mathcal{X}_{T_o} = \{X_{T_o}^i\}_{i=1}^N$ over the observation period $T_o = \{t_1, t_2, ..., t_o\}$, the trajectory prediction task is to find a set of their possible future trajectories $\mathcal{Y}_{T_p} = \{\hat{Y}^i_{T_p}\}_{i=1}^N$ during the prediction period $T_p = \{t_o + 1, t_o + 2, ..., t_o + t_p\}$.
Here, $X^i_{T_o} = \left\{(x^i_t, y^i_t)\right\}_{t=t_1}^{t_o}$ and $\hat{Y}^i_{T_p} = \left\{(\hat{x}^i_t, \hat{y}^i_t)\right\}_{t=t_o+1}^{t_o+t_p}$ represents agent $i$'s historical trajectory and future trajectory to be predicted respectively.
Subscripts indicating time or period (i.e., ${T_o}, t$) will be omitted for a clear representation.

Besides, let $I_t$ denote the scene visual image at time step $t$, and $({p_x}^i_t, {p_y}^i_t)$ be the position of agent $i$ at time $t$ in pixels.
Let $(x, y)$ represent agents' positions in \textbf{real} scales (like meters), $(p_x, p_y)$ represent them in \textbf{pixels}, and $(g_x, g_y)$ in \textbf{grids}.
Denote the mapping function from real-world positions to the pixel-positions as $(p_x, p_y) = m(x, y)$.

\subsection{Overview}
CSCNet contains three main parts:
An LSTM-based trajectory encoder to extract agents' historical movement patterns;
A context-aware transfer module that transfers variant scene visual images and interactive trajectories into the contextual domain to obtain a domain-invariant social-and-physical representation for each agent;
Finally, a trajectory decoder predicts the final positions with the condition of these transferred contextual images. 
See details in \FIG{fig_model}.

\subsection{Context-Aware Transfer}
We want our model to focus more on modeling the common parts of both social and physical interactions in different scenes rather than their rich appearance details.
To achieve this we use two separate modules to transfer interactive trajectories and scene images to a target domain that pays more attention to their activity semantics.

\subsubsection{Physical Transfer}
We wish to model physical constraints in a gentle way, rather than the hard semantic labels.
A natural thought is to infer the possibility of activities happening in some area according to its appearance.
We transfer the visual images to emphasize agents' activity semantics to build connections between trajectories and scenes, thereby better modeling the physical constraints in different scenes. 

When training on crowded datasets, we use all available trajectories in one scene as the supervision.
We first use the kernel density estimation method (KDE) to get the trajectory probability density (in \textbf{pixels}) in one certain dataset, i.e.,
\begin{equation}
    p(x, y) = \frac{1}{Rh^2}\sum_{i=1}^R K\left(\frac{x-{p_x}^i}{h}, \frac{y-{p_y}^i}{h}\right),
\end{equation}
where $R$ is the number of recorded positions (in pixels) in the dataset, $K$ is the kernel function and $h$ is its bandwidth.
To speed up calculation in experiments, we choose $K(x, y) = \max\left\{\frac{1}{h}-\frac{1}{h^2}\sqrt{x^2+y^2}, 0\right\}$.

KDE provides the ``pixel-level'' probability distribution of agents' possible active areas in the prediction scenes.
On one side, agents always synthesize and analyze scene information within a certain range of areas before making decisions.
The use of pixel-level probability distributions is not sufficient to fully reflect their future choices.
On the other side, when the number of agents available for estimating the active areas is insufficient, pixel-level activity labels will show significant quantization errors.
This will lead to label contamination in the training sample and further affect the model's accuracy to infer activity semantics from scene images.
Therefore, we collect the pixel-level active information in a certain range, and then construct a grid, i.e., a low-resolution histogram, to obtain a more robust behavioral representation.

We grid the scene images via the mapping function $(g_x, g_y) = m_{p \to g}(p_x, p_y)$ into the $H \times W$ grids to get the activity-semantic supervision label $l[I(g_x, g_y)]$ of each grid, i.e.,
\begin{equation}
    l[I(g_x, g_y)] = \iint_D p(x,y) \mathrm{d}x\mathrm{d}y,
\end{equation}
where $D = \left\{(x, y)| m_{p \to g}(x, y) = (g_x, g_y)\right\}$.
Thus, agents' activity records can be bound to the one grid visual image $I(g_x, g_y)$ through the active label $l[I(g_x, g_y)]$ in crowded scenes when training.

We employ a trainable transfer CNN (marked with $G(\cdot)$) to produce the corresponding grid $\hat{T} \in \mathbb{R}^{H \times W}$, i.e.,
\begin{equation}
    \hat{T} = G(I).
\end{equation}
We minimize the pixel-level error between the labels $l[I(g_x, g_y)]$ and predictions $\hat{T}(g_x, g_y)$.
It enables the network to directly predict the activity-semantics $\hat{\mathcal{T}} = \left\{\hat{T}^i\right\}$ only relaying on the scene images $\mathcal{I}$ during testing.

Moreover, the transfer CNN can forecast agents' possible active areas via scene images in different prediction scenes.
It means that the transferred images could adapt to different unseen scenarios, thus helping cross the Gap within the modeling of physical cues.

\subsubsection{Social Transfer}
The social transfer aims to represent complex social interactive behaviors through the energy map form, consistent with the above-transferred images describing physical constraints.
It makes it easier to analyze agents' movement trends and eliminate the gaps, which is caused by the different interactive preferences that may exist in different scenarios.

Given an empty grid $E^i \in \mathbb{R}^{H \times W}$, agent $i$'s historical trajectory $X^i$, and the set of his neighbors' (marked with $J$) trajectories $\mathcal{X}^{/i} = \left\{X^j | j \in J, j \neq i\right\}$, we define agent $i$'s interactive energy $E^i$ and the social transfer function $f$ as:
\begin{equation}
\begin{aligned}
    E^i &= f\left(X^i, \mathcal{X}^{/i}\right) = \lambda_1 E_{i \leftrightarrow i}^i + \lambda_2 E_{i \leftrightarrow j}^{i, J} + \lambda_3 E_{j \leftrightarrow j}^{i, J} \\
    &= \lambda_1 \sum_{(g_x, g_y) \in P_0(X^i)} -f\left(\frac{x-g_x}{h_1}, \frac{y-g_y}{h_1}\right) \\
    &+ \lambda_2 \sum_{X^j \in \mathcal{X}^{/i}} \sum_{(g_x^j, g_y^j) \in P_0(X^j)} - \theta_{ij} f\left(\frac{x-g_x^j}{h_2}, \frac{y-g_y^j}{h_2}\right) \\
    &+ \lambda_3 \sum_{X^j \in \mathcal{X}^{/i}} \sum_{(g_x^j, g_y^j) \in P_0(X^j)} f\left(\frac{x-g_x^j}{h_3}, \frac{y-g_y^j}{h_3}\right),
\end{aligned}
\end{equation}
where 
    $f(x, y) = \max\left\{1 - \sqrt{x^2+y^2}, 0\right\}$ is the base energy function,
    $P_0$ is the prior predictor (outputs in \textbf{grids}),
    $\theta_{ij}$ is the relative energy gain between agent $i$ and $j$,
    $\{h_1, h_2, h_3\}$ indicates the bandwidth of the energy function $f$,
    and $\{\lambda_1, \lambda_2, \lambda_3\}$ is a set of balance parameter.

The interactive energy is the linear combination of the base energy function with different scales:
The first item $E_{i \leftrightarrow i}^i$ shows agent $i$'s original future intention with the priori predictor $P_0$.
The second item $E_{i \leftrightarrow j}^{i, J}$ describes the intervention of others' activities on agents' original plannings.
Let $\vec{X^i} = (x^i_{t_o} - x^i_{t_1}, y^i_{t_o} - y^i_{t_1})^T$ denotes the moving vector of trajectory $X^i$, the relative energy gain is defined as:
\begin{equation}
    \theta_{ij} = \underbrace{\frac{\vec{X^i} \cdot \vec{X^j}}{\Vert\vec{X^i}\Vert_2 \Vert\vec{X^j}\Vert_2}}_{\mbox{direction gain}} \cdot \underbrace{\frac{\Vert\vec{X^i}\Vert_2}{\Vert\vec{X^j}\Vert_2}}_{\mbox{speed gain}} = \frac{\vec{X^i} \cdot \vec{X^j}}{\Vert\vec{X^j}\Vert_2^2}.
\end{equation}
It shows the interaction difference between agent $i$ and $j$ caused by the different motion directions and speed.
The last item $E_{j \leftrightarrow j}^{i, J}$ shows how social etiquette (i.e., the safety social distance) restricts agent $i$'s behavior.

Thus, we have transferred interactions from trajectories $\mathcal{X}$ into the energy view $\mathcal{E} = \left\{E^i\right\}$.
Note that all coordinates are calculated in the same grid scales as the physical transfer.
Loss functions used to tune these parameters can be seen in section ``Loss Functions''.

\subsection{Context Conditioned Prediction}
\paragraph{Context Representation}
The physical transfer transfers various scene images into the domain-invariant activity-semantic representations as $\hat{\mathcal{T}} = \left\{\hat{T}^i\right\}$.
The social transfer describes different social interactive behaviors in the energy view as $\mathcal{E} = \left\{E^i\right\}$.
Define the set $\mathcal{C} = \left\{C^i\right\}$ as the transferred images, which is the fusion of the above physical and social cues.
Each image $C^i$ is defined as:
\begin{equation}
    C^i = 1 - \frac{\hat{T}^i - \min \hat{\mathcal{T}} }{\max \hat{\mathcal{T}} - \min \hat{\mathcal{T}}} + \frac{E^i - \min \mathcal{E} }{\max \mathcal{E} - \min \mathcal{E}}.
\end{equation}
Then we employ a CNN to obtain context representations as:
\begin{equation}
    R_c^i = \mbox{CNN}(C^i).
\end{equation}

\paragraph{Historical Representation}
Like most previous methods \cite{socialGAN,liang2020temporal}, we guide LSTM as our feature extractor for agents' historical trajectories.
Agents' historical representations $\{R_h^i\}$ are obtained by:
\begin{equation}
    R_h^i = \mbox{LSTM}(X^i).
\end{equation}

\paragraph{Context Conditioned Prediction}
Using agent $i$'s historical representation $R_h^i$ and the context representation $R_c^i$, we employ the context conditioned decoder $D$ to predict agents' future positions, i.e.,
\begin{equation}
    \hat{Y}^i = D\left(\left[R_h^i, R_c^i\right]\right),
\end{equation}
where $[\cdot, \cdot]$ indicates the concatenating operation.
Note that the predictions are in real scales.
Please see the detailed structure of the above CNN and $D$ in Implementation Details.

\subsection{Loss Functions}
To train the whole network end to end, we use loss function:
\begin{equation}
\begin{aligned}
    \mathcal{L}_{\mbox{CSCNet}} &= \mu_1 \underbrace{\sum_i \Vert Y^i - \hat{Y}^i \Vert_2^2}_{\mbox{ADL}}
    + \mu_2 \underbrace{\sum_i \Vert T^i - \hat{T}^i \Vert_2^2}_{\mbox{STL}} \\
    &+ \mu_3 \underbrace{\sum_i \sum_{(\hat{x}^i, \hat{y}^i) \in \hat{Y}^i} C^i\left[m_{p \to g} (m (\hat{x}^i, \hat{y}^i))\right]}_{\mbox{CCL}},
\end{aligned}
\end{equation}
where $\{\mu_1, \mu_2, \mu_3\}$ is a set of balance parameters.

\paragraph{Average Displacement Loss}
The Average Displacement Loss (ADL) trains the entire model to generate reasonable trajectories similar to the ground truth.

\paragraph{Supervised Transfer Loss}
The Supervised Transfer Loss (STL) trains the physical transfer network $G(\cdot)$ to make it available to obtain the activity-semantics $\hat{T}$ from the scene images $I$ by minimizing the pixel-level error between $\hat{T}$ and the activity-semantic labels $T$.

\paragraph{Conditional Context Loss}
The Conditional Context Loss (CCL) trains the entire model to make the predictions under the condition of contextual transferred images.
It makes sure that all predictions have lower context energy.
The predicted areas have a higher possibility for all agents to act and lower interactive energy for the target agent to interact with others.

These three loss functions work together to make the whole network own the ability to predict agents' future positions under the constraints of social and physical rules.

%% file: content/fig_model.tex
\begin{figure*}[t]
\centering
\includegraphics[width=1\columnwidth]{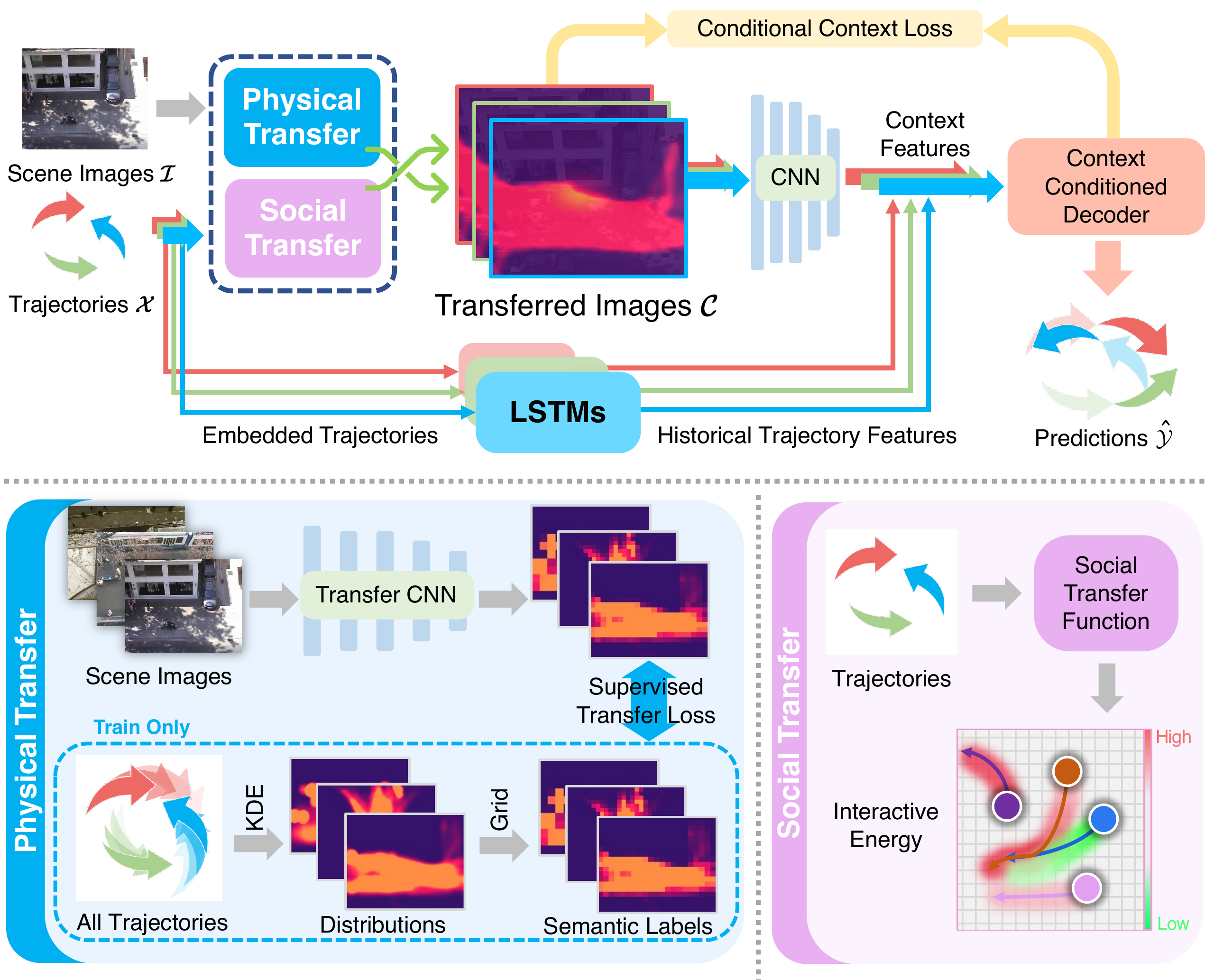}
\caption{
    The architecture of CSCNet.
    CSCNet uses agents' historical trajectories $\mathcal{X}$ and visual scene images $\mathcal{I}$ as the input to predict their reasonable future trajectories $\hat{\mathcal{Y}}$.
    We employ a context-aware transfer module to represent social cues and physical cues in a transferred image $\mathcal{C}$.
    Then the prediction network takes the $\mathcal{C}$ and $\mathcal{X}$ as the input to predict the future trajectories $\hat{\mathcal{Y}}$.
}
\label{fig_model}
\end{figure*}

%% file: content/Experiments.tex
\section{Experiments}
\subsection{Setup}
\paragraph{\textbf{Datasets}}
We choose three datasets, ETH-UCY and Stanford Drone Dataset (SDD), to evaluate our model.
These datasets cover challenging social behaviors such as walking together, groups crossing each other, following, etc.
\begin{itemize}
    \item ETH \cite{youWillNeverWalkAlone} - UCY \cite{2007Crowds} datasets consists of 1536 human trajectories in 5 scenarios: eth, hotel, zara1, zara2, and univ.
    In this case, agents' positions are labeled in meters.
    Following the experimental setup in \cite{socialLSTM}, we train models by the leave-one-out approach, which means that we train on all other sets and test on the remaining set.

    \item Stanford Drone Dataset (SDD) \cite{learningSocialEtiquette} collects videos captured from 20 unique scenes, which contains various types of targets (not just pedestrians, but also bikers, skateboarders, cars, buses, golf carts) that navigate in a university campus.
    Different from ETH-UCY, annotations are positions marked in pixels.
    Dataset splits used to train and validate are the same as Multiverse \cite{Liang_2020_CVPR}.
\end{itemize}

\paragraph{\textbf{Metrics}}
We use two widely used metrics, Average Displacement Error (ADE) and Final Displacement Error (FDE) \cite{socialLSTM}, to evaluate models' performance.
ADE is the average $l_2$ distance between predictions and the ground truth, and FDE is the $l_2$ distance of the last point's prediction and the ground truth.
Formally,
\begin{equation}
\begin{aligned}
    \mbox{ADE} &= \frac{1}{Nt_p} \sum_i \sum_t \left\Vert ({x}^i_t, {y}^i_t) - (\hat{x}^i_t, \hat{y}^i_t) \right\Vert_2,\\
    \mbox{FDE} &= \frac{1}{N} \left.\sum_i \left\Vert ({x}^i_t, {y}^i_t) - (\hat{x}^i_t, \hat{y}^i_t) \right\Vert_2 \right|_{t=t_o+t_p}.
\end{aligned}
\end{equation}
Besides, we choose minADE and minFDE \cite{stgcnn} among $K$ trajectories, i.e., ``best-of-$K$'', to evaluate stochastic models.

\subfile{./tb_ethucy.tex}

\paragraph{\textbf{Baselines}}
We choose 15 predictors as baselines.
They contain both deterministic models and stochastic models, which are as follows.
\subparagraph{Deterministic models} always predict $K=1$ trajectory, including 
\begin{itemize}
    \item S-LSTM \cite{socialLSTM} is one of the pioneering and classical models in trajectory prediction.
    \item Car-Net \cite{Sadeghian_2018_ECCV} is a deep attention-based model that combines two attention mechanisms for path prediction.
    \item SR-LSTM \cite{srLSTM} employs a data-driven states refinement LSTM network to enable the utilization of the current intention of neighbors through a message passing framework.
    \item STAR \cite{yu2020spatio} utilizes a spatio-temporal graph transformer framework, which tackles trajectory prediction by only attention mechanisms.
    \item E-SR-LSTM \cite{9261113} is the enhanced version of SR-LSTM \cite{srLSTM}, which applies spatial-edge LSTMs to exploit complementary features to SR-LSTM.
\end{itemize}
\subparagraph{Stochastic models} predict multiple trajectories ($K = 20$), including
\begin{itemize}
    \item S-GAN \cite{socialGAN} predicts socially plausible futures by training adversarially against a recurrent discriminator and encourages diverse predictions with a novel variety loss.
    We choose Social GAN-P containing the social pooling scheme as the baseline.
    \item NEXT \cite{peekingIntoTheFuture} is an end-to-end, multitask learning system utilizing rich visual features about the human behavioral information and interaction with their surroundings.
    \item SoPhie \cite{sophie}uses scene context information jointly with social interactions between the agents in order to obey physical constraints of the environment.
    \item Trajectron \cite{trajectron}is a graph-structured model that predicts many potential future trajectories of multiple agents simultaneously in both highly dynamic and multi-modal scenarios.
    \item S-BiGAT \cite{bigat} presents a graph-based
    A generative adversarial network that generates realistic, multi-modal trajectory predictions by better modeling the social interactions of pedestrians in a scene.
    \item STGCNN \cite{stgcnn} models the pedestrians' trajectories from the start as a spatio-temporal graph and proposes a weighted adjacency matrix in which the kernel function quantitatively measures the influence between pedestrians.
    \item Multiverse \cite{Liang_2020_CVPR} generates multiple plausible future trajectories, which contains novel designs of using multi-scale location encodings and convolutional RNNs over graphs.
    \item PECNet \cite{mangalam2020not} introduces a socially compliant, endpoint conditioned variational auto-encoder that closely imitates the multi-modal human trajectory planning process.
    \item TPNMS \cite{liang2020temporal} is the concurrent work, which builds a feature pyramid with increasingly richer temporal information to capture the motion behavior at various tempos better.
    \item SSALVM \cite{9160982} relies on contextual information that influences the trajectory of pedestrians to encode human-contextual interaction and model the uncertainty about future trajectories via latent variational mode.
\end{itemize}

\paragraph{\textbf{Implementation Details}}
Consistent with the most trajectory prediction approaches, we utilize agents' observed 8 frames of trajectories to predict their future 12 frames' trajectories, which is called the ``8-12-value protocol''.
The sample rate is 2.5 frames per second.
In addition, we use loss functions $\mathcal{L}_{\mbox{ADL}}$ and $\mathcal{L}_{\mbox{CCL}}$ to train the whole network' trainable variables, and $\mathcal{L}_{\mbox{STL}}$ to train the trainable variables in the transfer CNN ($G$) only.

In physical transfer, we set $h = 0.7$ meters to obtain the trajectories' possibility density.
The grid size is $(H, W) = (100, 100)$.
The transfer CNN ($G$) employs VGG19 pre-trained on ImageNet as the backbone, and we specify its output shape into $(100, 100)$ to generate the grid image $T$.
In social transfer, we guide the linear predictor as the $P_0$.
Parameters in social transfer function i.e., $\{h_1, h_2, h_3\}$, are made trainable and will be optimized via the $\mathcal{L}_{\mbox{CCL}}$.
The balance parameters in social transfer function are set as $\{\lambda_1, \lambda_2, \lambda_3\} = \{0.4, 0.4, 0.2\}$.
The balance parameters in loss functions are set as $\{\mu_1, \mu_2, \mu_3\} = \{0.3, 0.3, 0.4\}$.

The structure of CNN that extracts the context features is a stack of these layers: \{3*3 Average-Pooling, 11*11*32 Conv, 3*3*32 Conv, 3*3 Average-Pooling, 3*3*12 Conv, Flatten\} with ReLU activation.
Then we use one FC layer with tanh activation to obtain the 64-dimension context feature.
For the LSTM to extract historical features, we use an MLP to embed trajectories into a 64-dimension vector and input it.
We use tanh as the activation function, and the output of LSTM is a 64-dimension vector.
The trajectory decoder $D$ is an MLP with 3 layers, and output units of these layers are 128, 64, 2.
ReLU activations are applied except for the final output layer in it.

\subfile{./tb_SDD.tex}

\subfile{./fig_vis.tex}

\subsection{Quantitative Analysis}

\paragraph{\textbf{ETH-UCY}}
\TABLE{tb_ethucy} shows the results on ETH-UCY.
The deterministic CSCNet achieves noticeable performance gains than the state-of-the-art deterministic methods STAR and E-SR-LSTM.
STAR has an average error of 0.41 on ADE and 0.87 on FDE, while our CSCNet improves up to 10\% on ADE (0.37 meters) and 5\% on FDE (0.79 meters).
CSCNet also demonstrates strong competitiveness even compared to the stochastic methods.
For instance, CSCNet achieves similar performance in just one deterministic output compared to the concurrent work TPNMS, which generates 20 predicted trajectories.

\paragraph{\textbf{Stanford Drone Dataset}}
\TABLE{tb_SDD} presents the results on SDD.
Car-Net is the state-of-the-art deterministic method, and PECNet is the state-of-the-art stochastic method on SDD at present.
CSCNet achieves superior performance compared to Car-Net on both ADE/FDE by a significant margin of 43\% and 48\%, respectively.
The stochastic methods improve the metrics by predicting multiple trajectories. 
Apparently, CSCNet outperforms previous models whatever they are deterministic or stochastic.
When predicting $K=1$ trajectory, the average ADE and FDE of PECNet($K=1$) are 17.29 and 35.12 in pixels, respectively.
In comparison, CSCNet decreases these errors by about 15\% on ADE and 23\% on FDE.
Even compared to stochastic models ($K=20$), like Multiverse, CSCNet has a robust competitive edge.
It demonstrates that our proposed method performs better on both ETH-UCY and SDD with only one predicted trajectory.

\subfile{./fig_multi.tex}

\subsection{Qualitative Analysis}
This section provides some omnipresent examples in life to show that CSCNet can successfully handle complex scenarios.
We also compare CSCNet's visualized results with SR-LSTM qualitatively.
As shown in \FIG{fig_vis}, the images in the blue box show SR-LSTM's results.
The corresponding results of CSCNet are listed below, and the remaining images show the predictions given by CSCNet in different situations.
To better understand CSCNet, prediction situations with complex interactive context are shown in \FIG{fig_vis_multi}, including passing (Row 1), moving forward together (Row 2), avoiding the vehicle (Row 3).
It is well known that the real world changes continuously.
We provide several prediction cases sampled with the same time interval to analyze CSCNet's ability to cope with dynamic scenes and interactive context.

\paragraph{\textbf{Passing}}
$P2$ starts moving from standing during Step 1 to Step 4.
$P2$ does not notice $P1$ who fall behind at Step 1 and Step 2.
Under this situation, \MODEL~predicts that $P2$ will safely move straight in the future.
\MODEL~gives quite different prediction to $P2$ for it considers that $P2$ may notice $P1$ at Step 3.
The prediction of $P1$ also satisfies the above case at Step 4.
$P1$ first walks straight and finally turns to the store.
The predictions show the similar motion change, i.e., from walking straight to entering the store.
Therefore, it shows that CSCNet is adept at modeling dynamic behaviors at various tempos.

\paragraph{\textbf{Moving forward together}}
$P3$ and $P4$ walk together and turn around the car.
Our model provides predictions with similar trends.
It is worth mentioning that at step 4, CSCNet's predictions indicate that these two pedestrians' locations will converge in the future.
This interesting group phenomenon reflects that CSCNet can respond well to the group with the same motion plannings.

\paragraph{\textbf{Avoiding the Vehicle}}
Observing $P5$ and $P6$, we find that their predicted trajectories vary dynamically with the changing context (the moving car) from Step 1 to 4.
CSCNet can provide adaptive predictions for cases with different interactive context: walk straight (Step 1), decelerate and turn right (Step 2), stop to stand still (Step 3), turn left and accelerate (Step 4).
The prediction adaptively changes according to the vehicle's driving track, which shows CSCNet has a solid ability to deal with physical interaction.
Besides, \MODEL~gives interesting predictions for these two pedestrians at Step 3.
When the car drives up to them, it is predicted that they will decelerate and stand closer to each other.
It is consistent with the specific interaction phenomenon that people usually tend to huddle together for warmth when facing dangers.
Thus, it demonstrates that CSCNet can effectively capture the dynamic context changes in complex scenarios.

\subsection{Ablation Study}

\subfile{./tb_ab.tex}

There exist several kinds of semantic deviations inner or between social and physical interactions, which are the focus of this paper.
We aim to eliminate differences between semantics of social and physical interactions, which we call the ``Gap'' (See details in Section ``Introduction'').
Therefore, we design the context-aware transfer, including social transfer and physical transfer to cross the Gap.
Additionally, we propose the conditional context loss to train CSCNet to give predictions that satisfy contextual constraints.
To verify the superiority of the two essential components, we conduct the corresponding ablation experiments.

\subparagraph{Quantitative Results}
There are a series of methods showing the effects of two distinct cues, social and physical.
To better understand how social and physical cues affect agents' plannings, we provide two CSCNet variations: CSCNet-a0 and CSCNet-a1.
To further verify how theses components work, we choose four representative models to analyze and demonstrate the comparison, i.e., V-LSTM \cite{lstm}, SR-LSTM \cite{srLSTM}, SoPhie \cite{sophie}, and S-BiGAT\cite{bigat}.
\begin{itemize}
    \item V-LSTM does not handle any social or physical interactions.
    \item SR-LSTM only handles social interaction through an attention mechanism.
    \item SoPhie handles social and physical interactions through attention mechanism.
    \item S-BiGAT handles social interaction through GAT and physical interaction through VGG19.
    \item CSCNet-a0 is CSCNet without physical transfer (PT) and supervised transfer loss (STL).
    Balance parameters in the loss function are set as $\{\mu_1, \mu_2, \mu_3\} = \{0.5, 0, 0.5\}$.
    \item CSCNet-a1 is CSCNet without conditional context loss (CCL).
    The corresponding balance parameters are set as $\{\mu_1, \mu_2, \mu_3\} = \{0.5, 0.5, 0\}$.
\end{itemize}
Quantitative comparisons are shown in \TABLE{tb_ab}.

Obviously, the social module makes a difference comparing SR-LSTM and V-LSTM.
Meanwhile, the physical module also helps to improve the performance of prediction comparing SR-LSTM and SoPhie/S-BiGAT.
Despite utilizing different techniques to deal with social and physical interactions, SoPhie and S-BiGAT achieve similar performance.

Please note that CSCNet-a0 employs a CNN to extract features from scene images like SoPhie/S-BiGAT.
However, PT is not used to align the semantics of social and physical interactions.
As shown in \TABLE{tb_ab}, CSCNet-a0 achieves an average ADE/FDE improvement of 8.3\%/11.0\% compared to S-BiGAT.
Accordingly, ST can deal with social interaction better.

In the previous methods, the features extracted to model social and physical cues are simply concatenated.
PT aligns the two different semantics to maintain contextual consistency.
Comparing CSCNet and CSCNet-a0, we find a significant improvement of 15.9\%/11.2\% due to the help of PT and STL.
\MODEL~addresses the effects of physical and social cues for agent trajectory estimation through aligning the semantics.

CSCNet-a1 is trained without CCL only, and the transferred images are still used.
CSCNet outperforms CSCNet-a1 by a margin of 7.5\%/4.8\%.
Besides, CCL trends all predictions to have the lower context energy.
\MODEL~are trained end-to-end using CCL that keep consistency of the prediction with the generated prior.
More details are provided in \FIG{fig_ab}.

\subparagraph{Qualitative Results}

\subfile{./fig_ab.tex}

\FIG{fig_ab} shows the visualized comparison of CSCNet and CSCNet-a0.
(a) shows the effect of the context-aware transfer on physical constraints.
The man concerned in (a) walks straight in the past, but the parking car in front will prevent him from moving forward.
CSCNet-a0's predictions (red stars) seem not to consider this constraint.
With both context-aware transfer and CCL, CSCNet's predictions (yellow stars) are more physically acceptable.
(c) shows the effect on social constraints.
The man concerned in (c) walks towards another group of pedestrians.
He might collide with these pedestrians in the future, similar to CSCNet-a0's predictions (red stars).
Transferred images have modeled interactive behaviors in the form of energy (i.e., \FIG{fig_ab} (d)) so that CSCNet gives the socially acceptable predictions (yellow stars).

CSCNet can not achieve the final performance without either of the above items.
Transferred images provide the condition for different prediction situations, with the help of alignment on both social and physical descriptions at the semantic level.
The multiple modules in the model are trained end-to-end using a multi-target loss function that penalizes average error, intermediate representation accuracy, and consistency of the prediction with the generated prior.

%% file: content/tb_ethucy.tex
\begin{table*}[t]
\centering
\resizebox{1\columnwidth}{!}{
\begin{tabular}{c|c|ccccc|c}
\hline
Type & Model & eth & hotel & zara1 & zara2 & univ & AVG \\
\hline
\multirow{8}{*}{\rotatebox{90}{Deterministic}}
 & S-LSTM \cite{socialLSTM} & 1.09/2.35 & 0.79/1.76 & 0.47/1.00 & 0.56/1.17 & 0.67/1.40 & 0.72/1.54 \\
 & SR-LSTM \cite{srLSTM} & 0.63/1.25 & 0.37/0.74 & 0.41/0.90 & 0.32/0.70 & 0.51/1.10 & 0.45/0.94 \\
 & S-BiGAT(1) \cite{bigat} & N/A & N/A & N/A & N/A & N/A & 0.61/1.33 \\
 & E-SR-LSTM \cite{9261113} & 0.58/1.13 & 0.31/0.62 & 0.41/0.90 & 0.33/0.73 & 0.50/1.10 & 0.43/0.89\\
 & SSALVM(1) \cite{9160982} & 0.68/1.25 & 0.33/0.60 & 0.32/0.69 & 0.40/0.85 & 0.60/1.28 & 0.46/0.93\\
 & STAR \cite{yu2020spatio} & 0.56/1.11 & 0.26/0.50 & 0.40/0.89 & \textbf{0.31}/0.71 & 0.52/1.13 & 0.41/0.87 \\
 & CSCNet (Ours) & \textbf{0.51/1.05} & \textbf{0.22/0.42} & \textbf{0.36/0.81} & \textbf{0.31/0.68} & \textbf{0.47/1.02} & \textbf{0.37/0.79}\\
\hline
\multirow{8}{*}{\rotatebox{90}{\makecell[c]{Stochastic \\ $K=20$ }}}
 & S-GAN \cite{socialGAN} & 0.87/1.62 & 0.67/1.37 & 0.35/0.68 & 0.42/0.84 & 0.76/1.52 & 0.61/1.21 \\
 & NEXT \cite{peekingIntoTheFuture} & 0.73/1.65 & 0.30/0.59 & 0.38/0.81 & 0.31/0.68 & 0.60/1.27 & 0.46/1.00 \\
 & SoPhie \cite{sophie} & 0.70/1.43 & 0.76/1.67 & 0.30/0.63 & 0.38/0.78 & 0.54/1.24 & 0.54/1.15\\
 & Trajectron \cite{trajectron} & 0.59/1.14 & 0.35/0.66 & 0.43/0.83 & 0.43/0.85 & 0.54/1.13 & 0.46/1.00 \\
 & S-BiGAT(20) \cite{bigat} & 0.69/1.29 & 0.49/1.01 & 0.30/0.62 & 0.36/0.75 & 0.55/1.32 & 0.48/1.00\\
 & STGCNN \cite{stgcnn} & 0.64/1.11 & 0.49/0.85 & 0.34/0.53 & 0.30/0.48 & 0.44/0.79 & 0.44/0.75 \\
 & SSALVM(20) \cite{9160982} & 0.61/1.09 & 0.28/0.51 & 0.30/0.64 & 0.37/0.78 & 0.59/1.24 & 0.43/0.85\\
 & TPNMS$^*$ \cite{liang2020temporal} & 0.52/0.89 & 0.22/0.39 & 0.35/0.70 & 0.27/0.56 & 0.55/1.13 & 0.38/0.73 \\
\hline
\end{tabular}
}
\caption{
Quantitative results on ETH-UCY dataset.
$*$ indicates concurrent work. 
Error metrics are ADE/FDE in meters, and lower is better.
$k$ in $(k)$ represents the number of output trajectories for stochastic models and their results are validated with the best-of-$k$ validation.
For example, results of S-BiGAT(1) are reported as S-BiGAT when generating $K=1$ trajectory for each agent.
``N/A'' means the corresponding article \cite{bigat} has not provided detailed results on each ETH-UCY sub-dataset.
}
\label{tb_ethucy}
\end{table*}

%% file: content/tb_SDD.tex
\begin{table}[t]
\footnotesize
\centering
\begin{tabular}{c|c}
\hline
Deterministic Model & ADE/FDE \\
\hline
Car-Net \cite{Sadeghian_2018_ECCV} & 25.72/51.80 \\
PECNet(1) \cite{mangalam2020not} & 17.29/35.12 \\
CSCNet (Ours) & \textbf{14.63/26.91} \\
\hline
\hline
Stochastic Model (best-of-20) & ADE/FDE\\
\hline
S-GAN \cite{socialGAN} & 27.25/41.44 \\
SoPhie \cite{sophie} & 16.27/29.38 \\
Multiverse \cite{Liang_2020_CVPR} & 14.78/27.09 \\
PECNet(20) \cite{mangalam2020not} & 9.96/15.88 \\
\hline
\end{tabular}
\caption{
Quantitative results on SDD. 
Error metrics are ADE/FDE in pixels.
The Lower the better.
$k$ in $(k)$ represents the number of output trajectories for stochastic models.
}
\label{tb_SDD}
\end{table}

%% file: content/fig_vis.tex
\begin{figure*}[t]
\centering
\includegraphics[width=1\columnwidth]{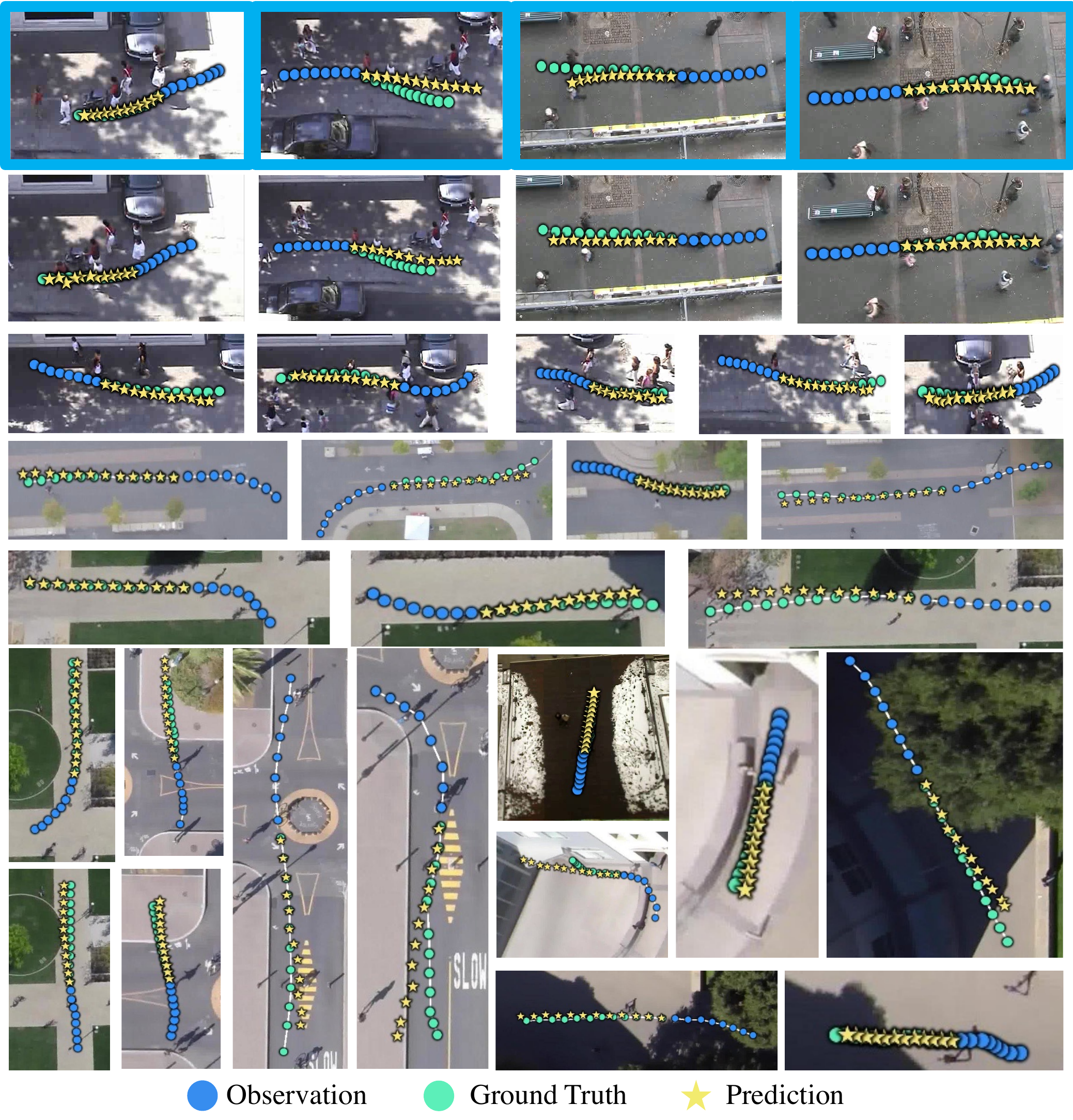}
\caption{
Visualized results of different models on ETH-UCY and SDD.
}
\label{fig_vis}
\end{figure*}

%% file: content/fig_multi.tex
\begin{figure*}[t]
\centering
\includegraphics[width=1\columnwidth]{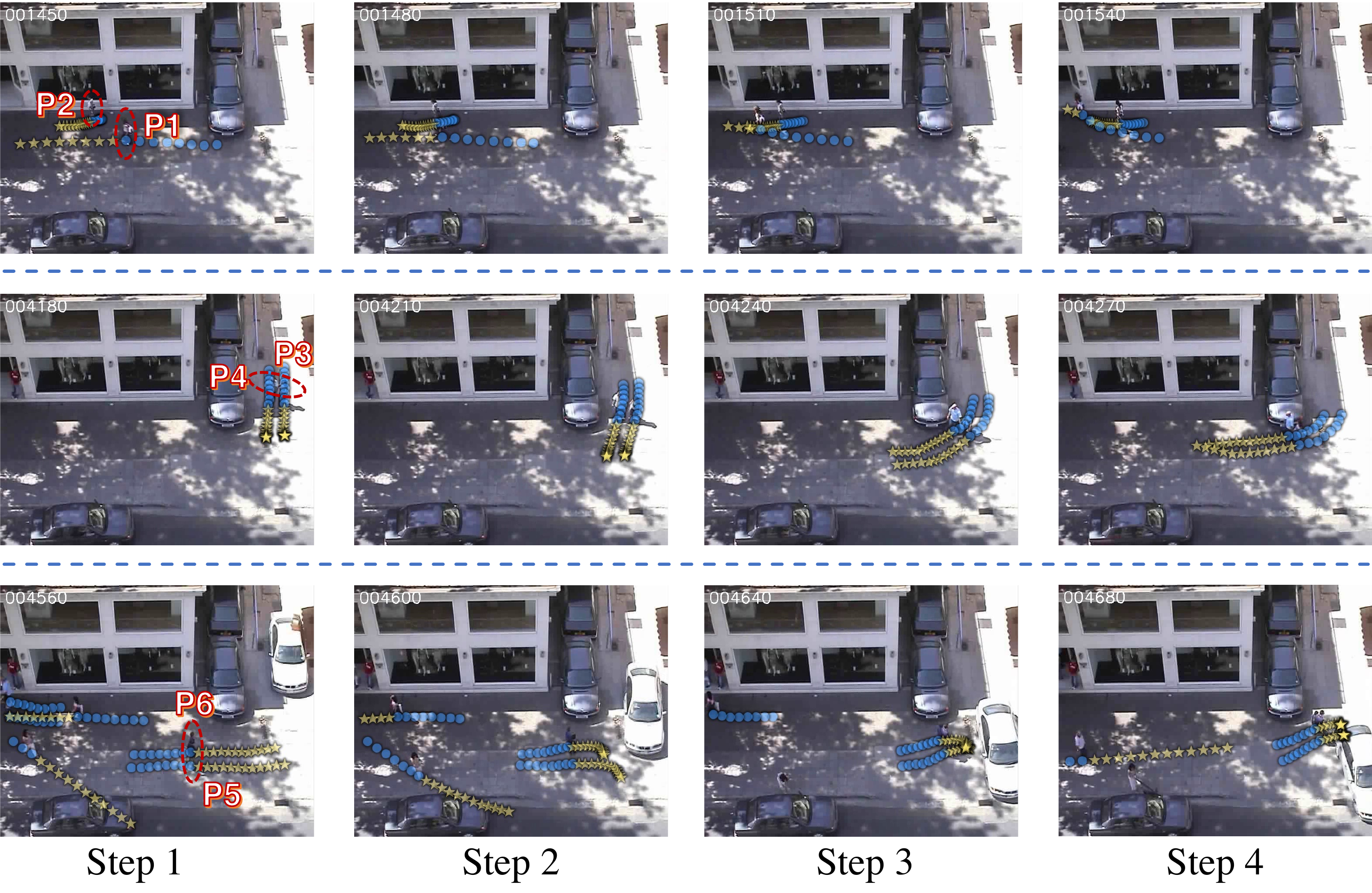}
\caption{
Visualized results of multiple pedestrians on UCY-zara1.
Row 1, Row 2, and Row 3 show the predictions under interactive behaviors like passing, moving forward together, and avoiding the vehicle, respectively.
Blue points represent agents' observations, and yellow stars represent the predicted trajectories.
Pedestrians marked by red dashed circles are the targets to be analyzed.
Predictions beyond the camera view are not displayed.
They are labeled as $P1 - P6$, respectively.
}
\label{fig_vis_multi}
\end{figure*}

%% file: content/tb_ab.tex
\begin{table*}[t]
\footnotesize
\centering
\begin{tabular}{c|c|c|c|c|c|c}
\hline
\multirow{2}{*}{Method} & \multicolumn{2}{c|}{Input Types} & \multicolumn{2}{c|}{Module} & \multirow{2}{*}{Ext. $\mathcal{L}$} & \multirow{2}{*}{ADE/FDE} \\
\cline{2-5} & TJ. & IMG. & SM & PM & \\
\hline
V-LSTM $^*$ & \checkmark & - & - & - & - & 0.70/1.52 \\
SR-LSTM \cite{srLSTM} & \checkmark & - & ATT. & - & - & 0.45/0.94 \\
SoPhie \cite{sophie} & \checkmark & \checkmark & ATT. & ATT. & - & 0.54/1.15 \\
S-BiGAT \cite{bigat} & \checkmark & \checkmark & GAT & VGG & - & 0.48/1.00 \\
\hline
CSCNet-a0 & \checkmark & \checkmark & ST & - & CCL & 0.44/0.89 \\
CSCNet-a1 & \checkmark & \checkmark & ST & PT & STL & 0.40/0.83 \\
CSCNet & \checkmark & \checkmark & ST & PT & STL \& CCL & 0.37/0.79 \\
\hline
\end{tabular}
\caption{
Ablation study on ETH-UCY datasets.
Error metrics are average ADE/FDE in meters.
The Lower the better.
``TJ.'' and ``IMG.'' indicate whether these models take agents' 2d coordinates and scene images as the input, respectively.
``SM'' and ``PM'' are short for modules that different models employ to model social cues and physical cues, respectively.
``Ext. $\mathcal{L}$'' indicates extra loss functions used in their models except the point-wise $l_2$ loss.
``ATT.'', ``GAT'', ``ST'', and ``PT'' indicate the attention mechanism, graph attention networks, social transfer, and physical transfer, respectively.
The results marked with $^*$ are directly obtained from \cite{stgat}.
}
\label{tb_ab}
\end{table*}

%% file: content/fig_ab.tex
\begin{figure*}[t]
\centering
\includegraphics[width=1\columnwidth]{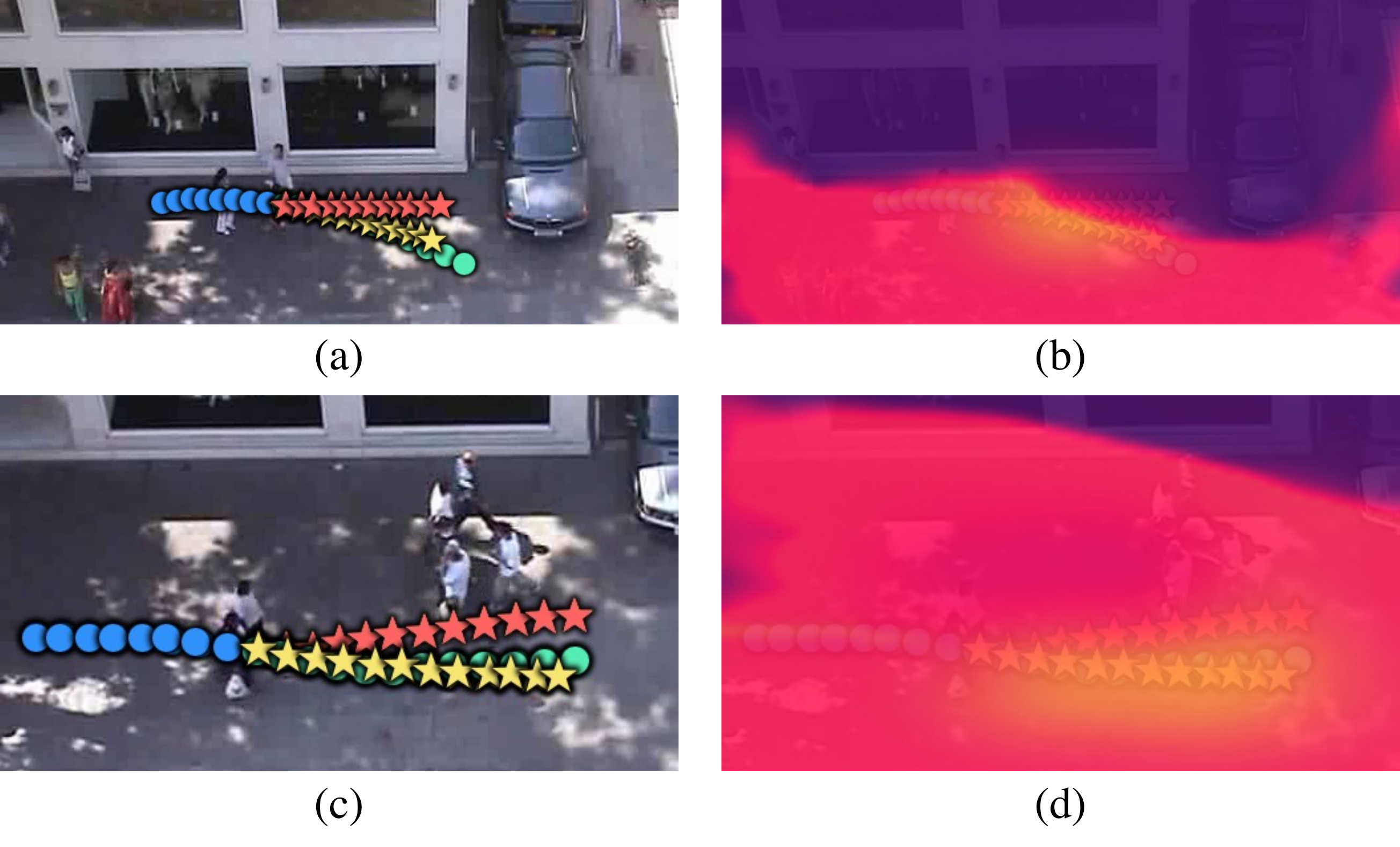}
\caption{
Qualitative ablation study.
(a) and (c) are two different prediction cases on UCY-zara1.
Blue points represent agents' observed trajectories, and green points are their ground truth future trajectories.
Yellow stars denote CSCNet's predictions, and red stars denote CSCNet-a0's (CSCNet-a0 is CSCNet without PT and STL).
(b) and (d) are corresponding transferred images of (a) and (c), which are covered on the scene images with a transparency of 0.9.
Yellow represents a lower value, and blue represents a higher.
The lower value means the predicted areas have a higher possibility for agents to pass through under social and physical constraints.
}
\label{fig_ab}
\end{figure*}

%% file: content/Discussion.tex
\section{Discussion}

\subfile{./fig_dis.tex}

\paragraph{\textbf{Error Analysis}}
We show some typical errors our model makes in \FIG{fig_dis}.
There are mainly two kinds of errors, including
\begin{itemize}
    \item Speed mismatch.
    Although the predicted direction is almost the same as its ground truth, the given speed does not match the reality in (a) - (c).
    CSCNet believes pedestrians may remain their previous speed if the interactive context has not changed drastically.
    \item State change.
    (d) - (f) show our model misses the correct direction since pedestrians change their motion states during the prediction period.
    For instance, the pedestrian in (e) is standing still.
    However, he stops talking to others and suddenly leaves the group in the future.
    Our proposed method does not consider the uncertainty of interactive factors, such as continuous conversation with others or the possibility of leaving the group.
\end{itemize}
In summary, the key reason is that CSCNet does not focus on the diversity of interaction situations when modeling interactions.
We will try to address these problems in our future works.

\paragraph{\textbf{Deterministic or Stochastic}}
More and more stochastic methods have been proposed since S-GAN \cite{socialGAN}.
These models provide agents with multiple future choices, which have been widely studied recently.
Stochastic models produce multiple future predictions by randomly sampling noisy variables, while discriminative models aim to produce a single future prediction to reflect the average choices among agents' future distributions.
Benefiting from the random sampling, stochastic models outperform deterministic models quantitatively when generating multiple predictions.
The goal of this work is to eliminate the discrepancy between the semantics of social and physical interactions.
As a deterministic model, CSCNet perhaps lacks the ability to describe agents' multiple plannings.
The proposed model uses a discriminative network with the LSTM as the backbone to give agents final predictions under the interactive context.
Experimental results demonstrate the performance improvement brought by the given context-aware transfer module.
In fact, our key design, i.e, the proposed transfer module, can be used for many forms of prediction networks, both stochastic and discriminative.
In future work, we will explore the use of context-aware transfer module in more prediction networks and further validate its effectiveness.

%% file: content/fig_dis.tex
\begin{figure*}[t]
\centering
\includegraphics[width=1\columnwidth]{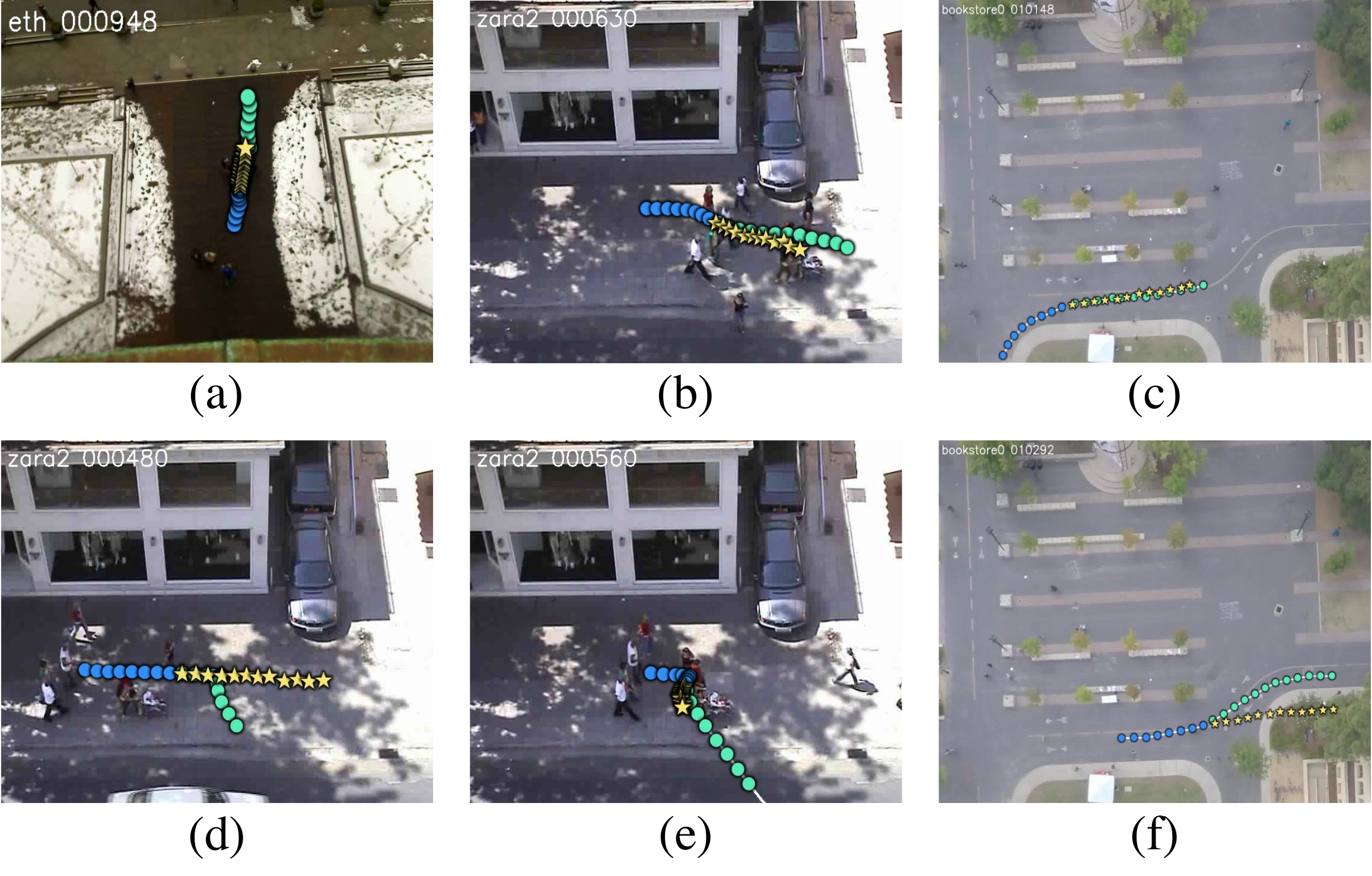}
\caption{
Typical error results on ETH-UCY and SDD.
(a) - (c) present ``Speed mismatch''.
(d) - (f) present ``State change''.
Blue points represent agents' observations, green points represent their ground truth future trajectories, and yellow stars represent the predicted trajectories.
}
\label{fig_dis}
\end{figure*}

%% file: content/Conclusion.tex
\section{Conclusion}
The modeling of interactive factors plays a significant role in the trajectory prediction task.
Many researchers are committed to this task by modeling factors like social interactions and physical interactions.
However, these interactive behaviors may show a shift in interaction semantics due to changes in prediction scenarios and agents' preferences.
Besides, the semantic shift phenomenon may also occur in different interactive representations due to various types of data formats.
In this paper, we propose CSCNet in a attempt to eliminate possible semantic differences between social and physical interactions when modeling agents' interactive context.
We design a semantic transfer module, the context-aware transfer, to align interactive representations on both social interactions and physical interactions at the semantic level with agents' realistic activities.
In detail, we introduce the intermediate representations, which represent physical cues and social cues with the same structures, to gather the common interactive preferences across multiple prediction scenes, thus reducing the impact brought by the semantic shift.
A transfer CNN and a social transfer function are proposed to align physical cues and social cues into the common transferred representations, respectively.
Then another context conditioned prediction sub-network is designed to give agents future predictions that are consistent with the interactive context.
The multiple modules will be trained end-to-end with a novel multi-target loss function, which penalizes average prediction error, intermediate representation accuracy, and the consistency of the prediction with the generated prior.
Finally, CSCNet outperforms 15 baseline models quantitatively and qualitatively on three public trajectory benchmark datasets, ETH-UCY and SDD.
Experimental results have shown the effectiveness of aligning social cues and physical cues at the semantic level when modeling these interactions across multiple datasets.
Additionally, we also analyze and discuss the limitations of the proposed model in several special interaction cases.
We will further solve these issues in our subsequent work and try to align more descriptions of agents' behaviors (e.g., activity labels, skeletons, etc.) through context-aware transfer to obtain more effective future prediction trajectories.

\section{Acknowledgement}
This work was supported partially by the National Natural Science Foundation of China (Grant No. 62172177, 62101179 and 61571205), in part by Natural Science Foundation of Hubei Province (Grant No. 2021CFB332), and in part by the Fundamental Research Funds for the Central Universities (Grant No. 2021yjsCXCY040, 2021yjsCXCY014 and 2021JYCXJJ037).

%% file: paper.bbl
\begin{thebibliography}{10}
\expandafter\ifx\csname url\endcsname\relax
  \def\url#1{\texttt{#1}}\fi
\expandafter\ifx\csname urlprefix\endcsname\relax\def\urlprefix{URL }\fi
\expandafter\ifx\csname href\endcsname\relax
  \def\href#1#2{#2} \def\path#1{#1}\fi

\bibitem{desire}
N.~Lee, W.~Choi, P.~Vernaza, C.~B. Choy, P.~H. Torr, M.~Chandraker, Desire:
  Distant future prediction in dynamic scenes with interacting agents, in:
  Proceedings of the IEEE Conference on Computer Vision and Pattern
  Recognition, 2017, pp. 336--345.

\bibitem{rhinehart2018r2p2}
N.~Rhinehart, K.~M. Kitani, P.~Vernaza, R2p2: A reparameterized pushforward
  policy for diverse, precise generative path forecasting, in: Proceedings of
  the European Conference on Computer Vision (ECCV), 2018, pp. 772--788.

\bibitem{rhinehart2019precog}
N.~Rhinehart, R.~McAllister, K.~Kitani, S.~Levine, Precog: Prediction
  conditioned on goals in visual multi-agent settings, in: Proceedings of the
  IEEE International Conference on Computer Vision, 2019, pp. 2821--2830.

\bibitem{unfreezing}
P.~Trautman, A.~Krause, Unfreezing the robot: Navigation in dense, interacting
  crowds, in: 2010 IEEE/RSJ International Conference on Intelligent Robots and
  Systems, IEEE, 2010, pp. 797--803.

\bibitem{learningSocialEtiquette}
A.~Robicquet, A.~Sadeghian, A.~Alahi, S.~Savarese, Learning social etiquette:
  Human trajectory understanding in crowded scenes, in: European conference on
  computer vision, Springer, 2016, pp. 549--565.

\bibitem{socialForce}
D.~Helbing, P.~Molnar, Social force model for pedestrian dynamics, Physical
  review E 51~(5) (1995) 4282.

\bibitem{socialLSTM}
A.~Alahi, K.~Goel, V.~Ramanathan, A.~Robicquet, L.~Fei-Fei, S.~Savarese, Social
  lstm: Human trajectory prediction in crowded spaces, in: Proceedings of the
  IEEE conference on computer vision and pattern recognition, 2016, pp.
  961--971.

\bibitem{ssLSTM}
H.~Xue, D.~Q. Huynh, M.~Reynolds, Ss-lstm: A hierarchical lstm model for
  pedestrian trajectory prediction, in: 2018 IEEE Winter Conference on
  Applications of Computer Vision (WACV), IEEE, 2018, pp. 1186--1194.

\bibitem{sophie}
A.~Sadeghian, V.~Kosaraju, A.~Sadeghian, N.~Hirose, H.~Rezatofighi,
  S.~Savarese, Sophie: An attentive gan for predicting paths compliant to
  social and physical constraints, in: Proceedings of the IEEE Conference on
  Computer Vision and Pattern Recognition, 2019, pp. 1349--1358.

\bibitem{bigat}
V.~Kosaraju, A.~Sadeghian, R.~Mart{\'\i}n-Mart{\'\i}n, I.~Reid, H.~Rezatofighi,
  S.~Savarese, Social-bigat: Multimodal trajectory forecasting using
  bicycle-gan and graph attention networks, in: Advances in Neural Information
  Processing Systems, 2019, pp. 137--146.

\bibitem{deo2018would}
N.~Deo, A.~Rangesh, M.~M. Trivedi, How would surround vehicles move? a unified
  framework for maneuver classification and motion prediction, IEEE
  Transactions on Intelligent Vehicles 3~(2) (2018) 129--140.

\bibitem{youWillNeverWalkAlone}
S.~Pellegrini, A.~Ess, K.~Schindler, L.~Van~Gool, You'll never walk alone:
  Modeling social behavior for multi-target tracking, in: 2009 IEEE 12th
  International Conference on Computer Vision, IEEE, 2009, pp. 261--268.

\bibitem{pei2019human}
Z.~Pei, X.~Qi, Y.~Zhang, M.~Ma, Y.-H. Yang, Human trajectory prediction in
  crowded scene using social-affinity long short-term memory, Pattern
  Recognition 93 (2019) 273--282.

\bibitem{barata2021sparse}
C.~Barata, J.~C. Nascimento, J.~M. Lemos, J.~S. Marques, Sparse motion fields
  for trajectory prediction, Pattern Recognition 110 (2021) 107631.

\bibitem{rossi2021human}
L.~Rossi, M.~Paolanti, R.~Pierdicca, E.~Frontoni, Human trajectory prediction
  and generation using lstm models and gans, Pattern Recognition 120 (2021)
  108136.

\bibitem{multiAgentTensorFusion}
T.~Zhao, Y.~Xu, M.~Monfort, W.~Choi, C.~Baker, Y.~Zhao, Y.~Wang, Y.~N. Wu,
  Multi-agent tensor fusion for contextual trajectory prediction, in:
  Proceedings of the IEEE Conference on Computer Vision and Pattern
  Recognition, 2019, pp. 12126--12134.

\bibitem{huang2021lstm}
Z.~Huang, J.~Wang, L.~Pi, X.~Song, L.~Yang, Lstm based trajectory prediction
  model for cyclist utilizing multiple interactions with environment, Pattern
  Recognition 112 (2021) 107800.

\bibitem{zamboni2022108252}
S.~Zamboni, Z.~T. Kefato, S.~Girdzijauskas, N.~Christoffer, L.~D. Col,
  Pedestrian trajectory prediction with convolutional neural networks, Pattern
  Recognition 121 (2022) 108252.

\bibitem{activityForecasting}
K.~M. Kitani, B.~D. Ziebart, J.~A. Bagnell, M.~Hebert, Activity forecasting,
  in: European Conference on Computer Vision, Springer, 2012, pp. 201--214.

\bibitem{cidnn}
Y.~Xu, Z.~Piao, S.~Gao, Encoding crowd interaction with deep neural network for
  pedestrian trajectory prediction, in: Proceedings of the IEEE Conference on
  Computer Vision and Pattern Recognition, 2018, pp. 5275--5284.

\bibitem{socialAttention}
A.~Vemula, K.~Muelling, J.~Oh, Social attention: Modeling attention in human
  crowds, in: 2018 IEEE international Conference on Robotics and Automation
  (ICRA), IEEE, 2018, pp. 1--7.

\bibitem{socialGAN}
A.~Gupta, J.~Johnson, L.~Fei-Fei, S.~Savarese, A.~Alahi, Social gan: Socially
  acceptable trajectories with generative adversarial networks, in: Proceedings
  of the IEEE Conference on Computer Vision and Pattern Recognition, 2018, pp.
  2255--2264.

\bibitem{stgat}
Y.~Huang, H.~Bi, Z.~Li, T.~Mao, Z.~Wang, Stgat: Modeling spatial-temporal
  interactions for human trajectory prediction, in: Proceedings of the IEEE
  International Conference on Computer Vision, 2019, pp. 6272--6281.

\bibitem{trafficPredict}
Y.~Ma, X.~Zhu, S.~Zhang, R.~Yang, W.~Wang, D.~Manocha, Trafficpredict:
  Trajectory prediction for heterogeneous traffic-agents, in: Proceedings of
  the AAAI Conference on Artificial Intelligence, Vol.~33, 2019, pp.
  6120--6127.

\bibitem{stgcnn}
A.~Mohamed, K.~Qian, M.~Elhoseiny, C.~Claudel, Social-stgcnn: A social
  spatio-temporal graph convolutional neural network for human trajectory
  prediction, in: Proceedings of the IEEE/CVF Conference on Computer Vision and
  Pattern Recognition, 2020, pp. 14424--14432.

\bibitem{peekingIntoTheFuture}
J.~Liang, L.~Jiang, J.~C. Niebles, A.~G. Hauptmann, L.~Fei-Fei, Peeking into
  the future: Predicting future person activities and locations in videos, in:
  Proceedings of the IEEE Conference on Computer Vision and Pattern
  Recognition, 2019, pp. 5725--5734.

\bibitem{Liang_2020_CVPR}
J.~Liang, L.~Jiang, K.~Murphy, T.~Yu, A.~Hauptmann, The garden of forking
  paths: Towards multi-future trajectory prediction, in: Proceedings of the
  IEEE/CVF Conference on Computer Vision and Pattern Recognition, 2020, pp.
  10508--10518.

\bibitem{Sun_2020_CVPR}
J.~Sun, Q.~Jiang, C.~Lu, Recursive social behavior graph for trajectory
  prediction, in: Proceedings of the IEEE/CVF Conference on Computer Vision and
  Pattern Recognition, 2020, pp. 660--669.

\bibitem{liang2020temporal}
R.~Liang, Y.~Li, X.~Li, J.~Zhou, W.~Zou, et~al., Temporal pyramid network for
  pedestrian trajectory prediction with multi-supervision, Proceedings of the
  AAAI conference on artificial intelligence (2021).

\bibitem{2007Crowds}
A.~Lerner, Y.~Chrysanthou, D.~Lischinski, Crowds by example, Computer Graphics
  Forum 26~(3) (2007) 655--664.

\bibitem{srLSTM}
P.~Zhang, W.~Ouyang, P.~Zhang, J.~Xue, N.~Zheng, Sr-lstm: State refinement for
  lstm towards pedestrian trajectory prediction, in: Proceedings of the IEEE
  Conference on Computer Vision and Pattern Recognition, 2019, pp.
  12085--12094.

\bibitem{9261113}
P.~Zhang, J.~Xue, P.~Zhang, N.~Zheng, W.~Ouyang, Social-aware pedestrian
  trajectory prediction via states refinement lstm, IEEE Transactions on
  Pattern Analysis and Machine Intelligence (2020) 1--1\href
  {https://doi.org/10.1109/TPAMI.2020.3038217}
  {\path{doi:10.1109/TPAMI.2020.3038217}}.

\bibitem{9160982}
A.~Díaz~Berenguer, M.~Alioscha-Perez, M.~C. Oveneke, H.~Sahli, Context-aware
  human trajectories prediction via latent variational model, IEEE Transactions
  on Circuits and Systems for Video Technology 31~(5) (2021) 1876--1889.
\newblock \href {https://doi.org/10.1109/TCSVT.2020.3014869}
  {\path{doi:10.1109/TCSVT.2020.3014869}}.

\bibitem{yu2020spatio}
C.~Yu, X.~Ma, J.~Ren, H.~Zhao, S.~Yi, Spatio-temporal graph transformer
  networks for pedestrian trajectory prediction, Proceedings of the European
  Conference on Computer Vision (ECCV) (2020) 507--523.

\bibitem{trajectron}
B.~Ivanovic, M.~Pavone, The trajectron: Probabilistic multi-agent trajectory
  modeling with dynamic spatiotemporal graphs, in: Proceedings of the IEEE
  International Conference on Computer Vision, 2019, pp. 2375--2384.

\bibitem{Sadeghian_2018_ECCV}
A.~Sadeghian, F.~Legros, M.~Voisin, R.~Vesel, A.~Alahi, S.~Savarese, Car-net:
  Clairvoyant attentive recurrent network, in: Proceedings of the European
  Conference on Computer Vision (ECCV), 2018.

\bibitem{mangalam2020not}
K.~Mangalam, H.~Girase, S.~Agarwal, K.-H. Lee, E.~Adeli, J.~Malik, A.~Gaidon,
  It is not the journey but the destination: Endpoint conditioned trajectory
  prediction, Proceedings of the European Conference on Computer Vision (ECCV)
  (2020) 759--776.

\bibitem{lstm}
S.~Hochreiter, J.~Schmidhuber, Long short-term memory, Neural computation 9~(8)
  (1997) 1735--1780.

\end{thebibliography}
